\documentclass[10pt, a4paper]{article}
\usepackage{lrec}
\usepackage{graphicx}
\usepackage{tabularx}
\usepackage{soul}
\usepackage{CJK}
\usepackage{booktabs}
\usepackage{color}
\usepackage{multirow}
\usepackage{epstopdf}
\usepackage[latin1]{inputenc}

\usepackage{hyperref}
\usepackage{xstring}

\title{The JDDC Corpus: A Large-Scale Multi-Turn Chinese Dialogue Dataset for E-commerce Customer Service}

\name{\parbox{0.7\textwidth}{\centering Meng Chen, Ruixue Liu, Lei Shen, Shaozu Yuan, Jingyan Zhou, Youzheng Wu, Xiaodong He, Bowen Zhou}}
\address{JD AI, Beijing, China \\
\{chenmeng20, liuruixue, shenlei41, yuanshaozu, zhoujingyan3\}@jd.com\\ \{wuyouzheng1, xiaodong.he, bowen.zhou\}@jd.com}

\abstract{Human conversations are complicated and building a human-like dialogue agent is an extremely challenging task. With the rapid development of deep learning techniques, data-driven models become more and more prevalent which need a huge amount of real conversation data. In this paper, we construct a large-scale real scenario Chinese E-commerce conversation corpus, \textbf{JDDC}, with more than 1 million multi-turn dialogues, 20 million utterances, and 150 million words. The dataset reflects several characteristics of human-human conversations, e.g., goal-driven, and long-term dependency among the context. It also covers various dialogue types including task-oriented, chitchat and question-answering. Extra intent information and three well-annotated challenge sets are also provided. Then, we evaluate several retrieval-based and generative models to provide basic benchmark performance on the JDDC corpus. And we hope JDDC can serve as an effective testbed and benefit the development of fundamental research in dialogue task.
 \\
\newline  }

\begin{document}

\maketitleabstract

\section{Introduction}
Building a human-like conversational agent is regarded as one of the most challenging tasks in Artificial Intelligence \cite{turing2009computing}. Because of the complexity, real scenario human conversations can be seen as a sequential, continuous, decision-making process, which relies on lots of information to make the conversation go on. For example, dialogue context, intents, external knowledge, common sense, emotions, participants' background and personas, etc. All these could have an impact on the response in a conversation. Moreover, these uncertainties make dialogue task extremely different from traditional machine learning tasks which usually have explicit targets and clearly defined evaluation metrics.      

To tackle this challenging problem, constructing a dialogue dataset is the most essential work. Especially for popular deep learning based approaches, large scale of training corpus in real scenario becomes decisive. However, existing datasets are still deficient. Datasets with structured annotations (e.g., slots and corresponding values) are often small-scale and in a limited capacity. Either traditional domain-specific ones \cite{allen1996robust,petukhova2014dbox,bordes2016learning,dodge2015evaluating} or recent multi-domain ones \cite{budzianowski2018multiwoz,shah2018building,el2017frames} are usually built for task-oriented dialogue systems. Another typical branch of works collects the dialogue corpus from movie subtitles, such as OpenSubtitles \cite{tiedemann2009news} and Cornell \cite{danescu2011chameleons}, which contain long sessions (over 100 turns) and some expressions like individual monologues may be not suitable for dialogue systems. More recently, some researchers construct dialogue datasets from social media networks (e.g., Twitter Dialogue Corpus \cite{Ritter2011Data} and Chinese Weibo dataset \cite{weibo13}), or online forums (e.g., Chinese Douban dataset \cite{Douban} and Ubuntu Dialogue Corpus \cite{Ubuntu}). Although in large scale, they are different from real scenario conversations, as posts and replies are informal, single-turn or short-term related. \\
In this work, we construct a large-scale multi-turn Chinese dialogue dataset, namely \textbf{JDDC} (Jing Dong Dialogue Corpus), with more than 1 million multi-turn dialogues, 20 million utterances, and 150 million words, which contains conversations about after-sales topics between users and customer service staffs in E-commerce scenario. Different from existing datasets mentioned above, the JDDC dataset illustrates the complexity of conversations in E-commerce. Table \ref{tab:example} presents a typical session in the corpus which contains services including: 1) task completion: changing the order address ($q_1$-$q_2$, text in blue); 2) knowledge-based Question Answering (QA): answering the question about refund period ($q_3$, text in red); and 3) feeling connection with the user: actively responding to the user's complains and soothe his/her emotion ($q_4$-$q_6$, text in purple). Therefore, this corpus supports to build a more challenging and comprehensive dialogue system. Additionally, the average conversation turn in a dialogue is 20, so the long-term dependency among the context is an important feature. As the example shown in Table \ref{tab:example}, to answer $q_4$, the assistant must look back to $r_2$ for further information. Besides, some contents in a real conversation are redundant or irrelevant to the final request. Taking $q_2$ as an example, the user explains the reason for his/her requirements, which actually has no contribution to solve the problem. Since all the data from our corpus is extracted from the real scenario, it reflects the characteristics in human-human conversations.
\begin{table*}[htb]
\small
\centering
\begin{tabular}{l|p{15cm}}
  \toprule[1pt]
  $q_1$ &  \begin{CJK*}{UTF8}{gbsn}\textcolor[rgb]{0,0,1}{可以帮我改下订单的地址吗？}\end{CJK*} \textcolor[rgb]{0,0,1}{(Could you help me change the address of the order?)} \\ \hline 
  $r_1$ &  \begin{CJK*}{UTF8}{gbsn}同一市内可以联系配送员直接修改的哦。 \end{CJK*} (You can contact the delivery staff directly if the two addresses are in the same city.)\\ \hline
  $q_2$ &  \begin{CJK*}{UTF8}{gbsn}\textcolor[rgb]{0,0,1}{不在同一个城市，现在地址是上海，但是我明天要回安徽。}\end{CJK*} \textcolor[rgb]{0,0,1}{(Not the same city. The current address is Shanghai, but I am going to Anhui tomorrow.)}\\ \hline
  $r_2$ &  \begin{CJK*}{UTF8}{gbsn}抱歉，地址在不同城市不能操作的，只能建议您重新下单哦。\end{CJK*} (Sorry, you cannot change the address to a different city. In this case, we suggest you place a new order.)\\ \hline
  $q_3$ &  \begin{CJK*}{UTF8}{gbsn}\textcolor[rgb]{1,0,0}{那我取消订单的话退款多久到账呢？}\end{CJK*} \textcolor[rgb]{1,0,0}{(How long does it take for the refund to arrive if I cancel the order?)}\\ \hline
  $r_3$ &  \begin{CJK*}{UTF8}{gbsn}微信零钱1个工作日内到账，储蓄卡1-7个工作日内到账，信用卡1-15个工作日内到账的哦！\end{CJK*} (For Wechat change, it  arrives in 1 working day. For debit card, it arrives in 1-7 working days. And for credit card, it arrives in 1-15 working days.)\\ \hline
  $q_4$ &  \begin{CJK*}{UTF8}{gbsn}\textcolor[rgb]{0.7,0.3,0.7}{为什么不能改地址，你们这也太不方便了。}\end{CJK*} \textcolor[rgb]{0.7,0.3,0.7}{(Why can't I change my address? That is too inconvenient.)}\\ \hline
  $r_4$ &  \begin{CJK*}{UTF8}{gbsn}非常抱歉，我们物流还有待完善呢。\end{CJK*} (I'm sorry. Our logistics system needs to be improved.)\\ \hline
  $q_5$ &  \begin{CJK*}{UTF8}{gbsn}\textcolor[rgb]{0.7,0.3,0.7}{这也太麻烦了，我还急着用呢。}\end{CJK*} \textcolor[rgb]{0.7,0.3,0.7}{(That is too troublesome, I'm in a hurry.)}\\ \hline
  $r_5$ &  \begin{CJK*}{UTF8}{gbsn}非常抱歉！如果是我的话我也会很着急的，我们会改进的！\end{CJK*} (I'm so sorry! If I were you, I would feel the same. We will do our best to improve it!)\\ \hline
  $q_6$ &  \begin{CJK*}{UTF8}{gbsn}\textcolor[rgb]{0.7,0.3,0.7}{行吧。}\end{CJK*} \textcolor[rgb]{0.7,0.3,0.7}{(Fine.)}\\ \hline
  $r_6$ &  \begin{CJK*}{UTF8}{gbsn}谢谢您的理解！还有什么能帮到您的吗？\end{CJK*} (Thanks for your understanding! What else can I do for you?)\\ 
  \bottomrule[1pt]
\end{tabular}
\caption{An example from JDDC corpus. The actual corpus is in Chinese. Best viewed in color.}
\label{tab:example}
\end{table*}\\
To bring the dataset more valuable for dialogue research, we label the intent for each query in all dialogues with a high-precision in-house classifier, which covers 289 different intents in real E-commerce after-sales scenario. We also prepare three Challenge Sets for evaluating dialogue systems better. In each set, different input information is provided and multiple ground-truth answers are annotated. We plan to annotate more information (e.g., emotions and external knowledge) for this dataset in the future. We hope the JDDC corpus can serve as an effective testbed for multi-turn dialogue research, to drive the development of fundamental techniques such as representation learning, neural-symbolic learning, reinforcement learning, knowledge-based reasoning, context modeling, and controllable response generation, etc.

In the following parts, related work is presented in Section \ref{label:related work}. Section \ref{label:data construction} illustrates the dataset construction process and detailed characteristics. We then evaluate existing mainstream approaches, including retrieval-based and generative approaches on the developed datasets in Section \ref{label:experiments}. Finally, Section \ref{label:conclusion} concludes the paper.
\begin{table*} 
\small
\centering
\begin{tabular}{p{3.5cm}<{\centering}|p{2cm}<{\centering}|p{2cm}<{\centering}|p{2cm}<{\centering}|p{5cm}<{\centering}}   
\toprule[1pt]
Dataset &Dialogues &Utterances & Words &Description\\ 
\hline
\multicolumn{1}{m{3cm}|}{\centering Twitter Corpus \cite{Twitter2010dataset}}&1,300,000&3,000,000&-& Post-reply pairs extracted from Twitter (English)\\
\hline
\multicolumn{1}{m{3cm}|}{\centering Weibo Corpus \cite{weibo15}} &4,435,959 &8,871,918&-& Post-comment pairs extracted from Weibo.com (Chinese) \\
\hline
\multicolumn{1}{m{3cm}|}{\centering Ubuntu Corpus \cite{Ubuntu}}& 930,000 & 7,100,000 & 100,000,000 & Post-reply chat logs from Ubuntu Forum (English) \\
\hline 
\multicolumn{1}{m{3cm}|}{\centering Douban Corpus \cite{Douban}}& 1,060,000
&7,092,000&131,747,880& Post-reply chat logs from Douban (Chinese)\\ 
\midrule[1pt]
\multicolumn{1}{m{3cm}|}{\centering PERSONA-CHAT \cite{facebook2018person}}&10,907&162,064&-& Personalizing chit-chat dialogue corpus (English)\\  
\hline 
\multicolumn{1}{m{3cm}|} {\centering DuConv Corpus \cite{DuConv}} &29,858 &270,399&2,872,340& Knowledge-driven conversation dataset (Chinese)\\  
\hline  
\multicolumn{1}{m{3cm}|} {\centering SGD Corpus \cite{rastogi2019towards}}&16,142 &659,928&3,217,149& Multi-domain task-oriented dialogue corpus (English)\\  
\midrule[1pt]
\multicolumn{1}{m{3cm}|} {\centering ECD Corpus \cite{Ali}}&1,020,000&7,500,000&49,000,000& E-commerce dialogue corpus from Taobao (Chinese)\\  
\hline  
\multicolumn{1}{m{3cm}|}{\centering \textbf{ \\ JDDC Corpus}} & 1,024,196 & 20,451,337 & 150,716,172 & E-commerce dialogue corpus from JD (Chinese) \\
\bottomrule[1pt]
\end{tabular} 
\caption{
Existing related large-scale datasets applicable to dialogue systems. Note that '-' represents the number is not mentioned in related papers.
} 
\label{related dataset}
\end{table*} 

\section{Related Work}
\label{label:related work}
The research on chatbots and dialogue systems has kept active for decades. The growth of this field has been consistently supported by the development of new datasets. We briefly review existing dialogue datasets, and roughly divide them into three categories according to data features: 1) large scale data extracted from social media or forums, 2) artificial dialogue corpus constructed from crowd workers, and 3) corpus collected from real human-human coversation scenario. A list of related large-scale datasets discussed is provided in Table \ref{related dataset}.

Traditional methods tend to extract conversation alike information from social media or forum \cite{Twitter2010dataset,weibo15,Douban,Ubuntu,LiA,Reddit}. Despite of the massive number of utterances included in these datasets, they usually provide ambiguous dialogue flows. It's due to the fact that these datasets mainly comprise post-reply pairs on social networks or forums where people interact with others more freely (often more than two speakers are involved in the conversation). Moreover, the replies in these datasets are most or only related to the post and there are very few context information provided for query understanding. \\
To imitate the natural conversation flows in real life, some datasets are collected with pre-defined prompts or guided schema. The DuConv \cite{DuConv} and PERSONA-CHAT \cite{facebook2018person} datasets are collected with Wizard-of-Oz technique \cite{WOZ}. The former one is collected during knowledge-driven conversation with one person playing as the conversation leader and the other one playing as the follower. However, the conversation goal is defined in advance. The later one collects data from two crowd workers with different persona information provided during conversation. Apart from the WOZ, the SGD \cite{rastogi2019towards} dataset is constructed by firstly generating dialogue outlines by simulator, then uses a crowd-sourcing procedure to paraphrase the outlines to natural language utterances. Even though these datasets keep the nature of conversation flow in some sense, offering large scale of conversation information is still infeasible. In that case DuConv only consists of less than 30k dialogues. SGD and PERSONA-CHAT datasets present even less dialogue information. \\
The most similar dataset to our JDDC corpus is ECD \cite{Ali} corpus which is also collected from real E-commerce scenario. Although it keeps the bi-turn information for real conversation and provides a considerable number of utterances, there are less turns offered for each conversation (7 turns per session) and no annotated test data is provided. Compared to ECD \cite{Ali}, the JDDC corpus has much longer context (average 20 turns per session), and we also prepare three high-quality human-annotated evaluation sets. Besides, extra intent information for each query is provided. These intents contain beneficial information for dialogue system to understand queries under complicated after-sales circumstances. 

\section{Dataset Construction}
\label{label:data construction}
\subsection{Data Collection and Statistics}
\begin{figure}[htb]
\begin{center}
   \includegraphics[width=1.0\linewidth]{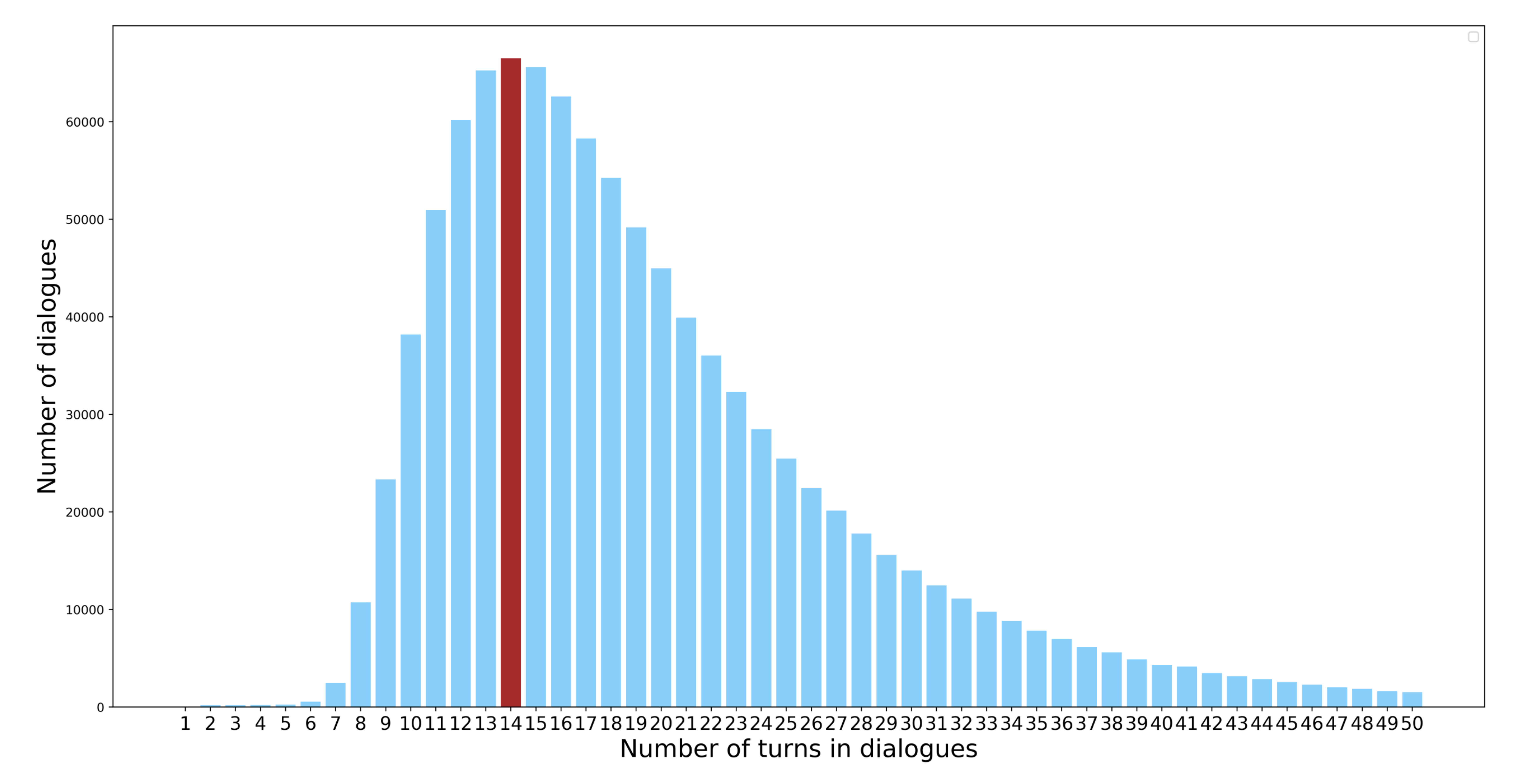}
\end{center}
   \caption{Histogram of dialogue turns in JDDC dataset.}
\label{fig:turns_dis}
\end{figure}

We collect the conversations between users and customer service staffs from Jing Dong (JD)\footnote{\url{https://www.jd.com}}, which is a popular E-commerce website in China. After crawling, we de-duplicated the raw data, desensitized and anonymized private information (e.g. replacing all numbers with special token $<$NUM$>$, and replacing order IDs with $<$ORDER-ID$>$). Then, we adopt Jieba\footnote{\url{https://github.com/fxsjy/jieba}} toolkit to perform Chinese word segmentation. We also count tokens, sessions and the average dialogue turns to give a brief view of the dataset. From Table \ref{tab:dataset_statistics}, we can see that the JDDC dataset contains 1,024,196 multi-turn sessions and 20,451,337 utterances totally. Besides, the number of turns for each session ranges from 2 to 83 with an average of 20. And the average tokens per utterance is about 7.4. Figure \ref{fig:turns_dis} demonstrates the histogram of dialogue lengths in the dataset. For space limitation, we only show the dialogues whose turns are less than 50. We can see that, most conversions are between 9 to 30 turns and sessions of 14 turns have the largest portion. This indicates that long-term dependency among context is a distinctive feature in the JDDC dataset.
\begin{table}[htb]
\small
\centering
\begin{tabular}{r c}
  \toprule[1pt]
  Total sessions & 1,024,196 \\
  Total utterance & 20,451,337 \\
  Total words & 150,716,172 \\
  Average words per utterance & 7.4 \\
        Average turns per session & 20 \\
    Max turns& 83 \\
    Min turns& 2 \\

  \bottomrule[1pt]
\end{tabular}
\caption{Basic statistics of JDDC dataset.}
\label{tab:dataset_statistics}
\end{table}
\subsection{Intent Distribution}
\begin{figure}[htb]
\begin{center}
   \includegraphics[width=1.0\linewidth]{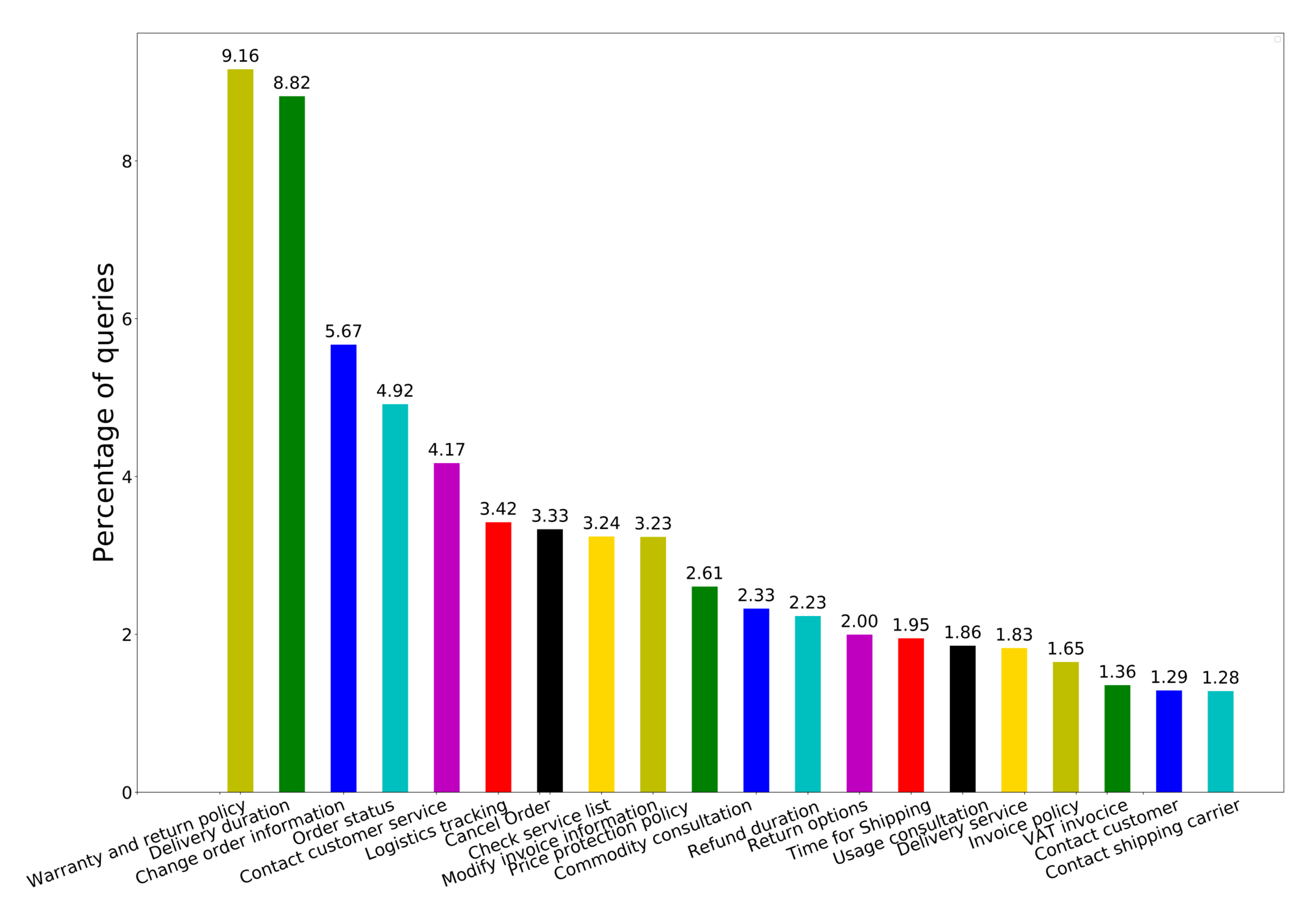}
\end{center}
   \caption{Distribution of intents in JDDC dataset.}
\label{fig:intent_dis}
\end{figure}

Different from open-domain chit-chat or task-oriented dialogue (e.g. booking restaurants or flight tickets), conversations in E-commerce after-sales scenario usually have explicit goals, which can be returning product, changing delivery address, or simply inquiring the warranty policy etc. So knowing the goals is important for modeling this dialogue task. To facilitate the research in the future, we label the intent for each query in all dialogues with a high-quality in-house intent classifier. The classifier contains totally 289 intents, and it's trained with Hierarchical Attention Network \cite{HAN} model so context is also considered. The training data for the classifier includes totally 578,127 instances and all the training instances are annotated by professional customer service staffs. The classification accuracy reaches 93\%, so the predicted intents for the JDDC dataset are reliable.\\
Figure \ref{fig:intent_dis} gives the distribution of top 20 intents. The top five intents are: `Warranty and return policy', `Delivery duration', `Change order information', `Check order status' and `Contact customer service'. Among them, `Warranty and return policy' accounts for 9.2\%, which is the most common intent in after-sales circumstance. This distribution is also consistent with our real experience in E-commerce scenario that people often concern about warranty and delivery cycle, and ask for changing or returning the product.
\subsection{Challenge Set}
\begin{figure}[htb]
\begin{center}
   \includegraphics[width=1.0\linewidth]{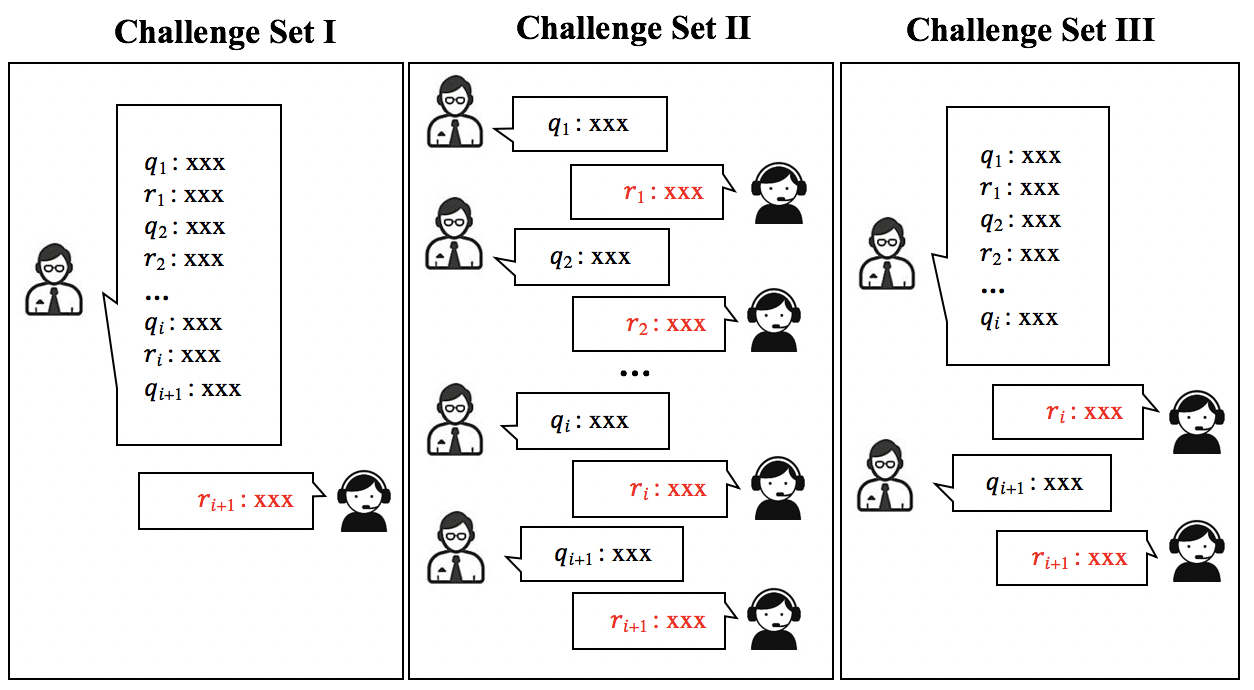}
\end{center}
   \caption{The explanation of our 3 challenge sets. The responses ($r$) in red color are required to be answered by the dialogue system.}
\label{fig:challenge}
\end{figure}

To promote the research of human-machine dialogue systems with massive data in real scenario, we also held large-scale multi-turn dialogue competition with the JDDC dataset, namely, JingDong Dialogue Chanllenge\footnote{\url{http://jddc.jd.com/}} in 2018 and 2019. Aiming to fully evaluate the dialogue systems submitted in the competitions, we released 3 challenge sets with different input information, and also annotated multiple ground-truth answers for each task. To further clarify these challenge sets, diagrams shown in Figure \ref{fig:challenge} illustrate the difference between 3 tasks.

\textbf{Challenge Set I}, the dialogue system is required to output the final response $r_{i+1}$ by utilizing multi-turn dialogue context in the format of $\{q_1, r_1, q_2, r_2, ... ,q_{i+1}\}$, where $q$ represents question and $r$ means response. This task is designed for long context modeling. Totally 300 dialogues are annotated, and 300 questions need to be answered in this set.

\textbf{Challenge Set II}, we mask the answers of a multi-turn conversation as $\{q_1, q_2,  ...,q_{i+1}\}$, and the dialogue system is required to generate answers $\{r_1,r_2,...,r_{i+1}\}$ according to sequential questions. Particularly, this requires considering not only the input question but also the generated responses in previous turns. This task is more challenging than the previous one because incorrect replies may mislead the next output. Totally 15 dialogues are annotated with 168 questions to be answered in this set.

\textbf{Challenge Set III}, we combine the characteristics of the former two tasks, for which the model needs to generate the response $\{r_{i},r_{i+1}\}$ sequentially under circumstance of several rounds of dialogue context and sequential questions $\{q_1,r_1,q_2,r_2,...,q_{i},q_{i+1}\}$. The questions here are mainly long-tailed and hard questions compared with Challenge Set I and II. Totally 108 dialogues are annotated and 500 questions need to be answered in this set.

In order to assess the answers generated by dialogue system, for each question, we annotated several candidate answers (10 candidate answers for Challenge Set I and II, and 3 for Challenge Set III), as for dialogue task, the ground-truth is usually not limited to one. What's more, different weights are provided for each candidate answer, so evaluation metrics (e.g., BLEU score) can be calculated more accurately. We hope these 3 challenge sets can help evaluate the dialogue systems on a fine-grained level.

\section{Experiments}
\label{label:experiments}
In this section, we conduct experiments on the JDDC dataset. We focus on two categories of models used in data-driven dialogue systems: retrieval-based models based on BM25 and BERT \cite{devlin2019bert} and generative models \cite{gu2016incorporating}. We will introduce some empirical settings, including dataset preparation, baseline methods, parameter settings. Then we introduce the experimental results on this dataset.

\subsection{Experimental Setup}
We first divide the around 1 million conversation sessions into training, validation and testing set. Then we construct $I$-$R$ pairs from each set into the $\{I, R\} = \{q_1, r_1, q_2, r_2, Q, R\}$ format, where $I = \{C, Q\}$ stands for input, $C = \{q_1, r_1, q_2, r_2\}$ is the dialogue context and $Q$ represents the last query. So the most recent two rounds of dialogue are kept as context. We also filtered some too short/long dialogues during experiment preparation. The statistics of the pre-processed dataset for experiment are shown in Table \ref{tab:experiment_statistics}. 

\begin{table}[htb]
\centering
\begin{tabular}{lccc}
  \toprule[1pt] 
   & Train & Valid & Test\\
  Sessions & 963,358 & 4,992 & 4,992\\
  I-R Pairs & 1,522,859 & 5,000 & 5,000\\
  \bottomrule[1pt]
\end{tabular}
\caption{JDDC dataset division in experiments.}
\label{tab:experiment_statistics}
\end{table}

For retrieval-based models, the original $I$-$R$ pairs in the training set are labelled as positive and negative responses are selected randomly from the dataset. The ratio of positive and negative sample is 1:1. Then the constructed positive and negative $I$-$R$ pairs are used to fine-tune the BERT model. Our model implementation for BERT is based on Google's work \cite{devlin2019bert} and follows the hyper-parameter settings in the original model.

For generative models, we first clean the training set to decrease the portion of short responses (shorter than 3 Chinese characters) and generic responses (e.g., ``What else can I do for you?''). Then all remaining $I$-$R$ pairs are used for training the model. Our code implementation is based on the machine translation toolkit OpenNMT \cite{klein2017opennmt}. In all generative  experiments, we set 100,000 for vocabulary size and 200 for word embedding dimension. The source length is 128 and target length is decreased to 40 to avoid generating too long response. Other training parameters are set as default.

\subsection{Comparable Models}
In this subsection, we will introduce the detailed information on retrieval-based models and generative models used for our experiment. 
\subsubsection{Retrieval-based Models}
\textbf{BM25}
To make the retrieval baseline more efficient, we firstly index all the Input-Response pairs in the training set using ElasticSearch\footnote{\url{https://www.elastic.co/products/elasticsearch}}. Then we use BM25 to retrieve the top 20 candidates for further matching. Response from top 1 candidate is used for evaluation. The equation of BM25 is defined as:
\begin{equation} \label{eq:bm25}
    S(I_{test}, I_{doc}) = \sum\limits_{i}^{n} W_i \cdot S_1(w_i, I_{doc}) \cdot S_2(w_i, I_{test})
\end{equation}
where $I_{test}$ stands for the test input including context $C$ and query $Q$, $w_i$ is the $i$-th word in the $I_{test}$, $I_{doc}$ is the document input in the repository, $W_i$ represents the weight of $w_i$ (such as inverse document frequency), and $S$($\cdot$) calculates the relevance score of the two elements. Therefore, $S(I_{test}, I_{doc})$ is the similarity score between the test input and the existing $I$-$R$ pairs in the repository.

\textbf{BERT-Retrieval}
The retrieval method above only use one lexical feature to calculate the similarity. To capture more semantic information, we fine-tune the pre-trained BERT model \cite{devlin2019bert} and add a dense layer with softmax as classifier to get the semantic similarity score for every ($I_{test}$, $R_{doc}$) pair. Then we use the BERT score to re-rank the top 20 candidates from ElasticSearch and return the final top 1.
\subsubsection{Generative Models}
\textbf{Vanilla Seq2Seq}
We implement the vanilla Sequence-to-Sequence (Seq2Seq) model \cite{seq2seq} with 512-unit 4-layer Bi-LSTMs for both the encoder and decoder. The input is the concatenated context and query, while the output is the response.

\textbf{Attention-based Seq2Seq}
To improve our baseline, we applied attention mechanism \cite{attention-luong} in the Seq2Seq model. This model is regarded as our second baseline and referred as Seq2Seq-Attention.

\textbf{Attention-based Seq2Seq with Copy}
The context-query input is usually long and contains a lot of rare terminologies like ``\begin{CJK*}{UTF8}{gbsn}京东白条\end{CJK*}'' (Jing Dong IOU(I owe you)), which may be OOV (out of vocabulary) words. Therefore, we add the copy mechanism \cite{gu2016incorporating} to the attention-based Seq2Seq baseline (Seq2Seq-Copy). The copy mechanism can explicitly extract words or phrases like certain entities from the input.

\subsection{Evaluation Measures}
In order to provide comparable baseline results for future research, we use some quantitative metrics for automatic evaluation. BLEU and ROUGE scores, which are widely used in NLP and multi-turn dialogue generation tasks \cite{tian2017make,luo2018auto,shen2019modeling}, are used to measure the quality of generated responses via the comparison with the ground truths. The recently proposed Distinct (Distinct-1/2) \cite{li2016diversity}, is used to evaluate the degree of diversity by calculating the ratio of unique unigrams and bigrams in the generated responses.

\subsection{Experimental Results}
In this section, we analyze different baselines' performance based on automatic evaluation measures and present in-depth case study. 

\subsubsection{Automatic Evaluation Results}

\begin{table}[htb]
\small
\centering
\begin{tabular}{p{1.5cm}<{\centering}|p{1.1cm}<{\centering}p{1.3cm}<{\centering}p{1.1cm}<{\centering}p{1.1cm}<{\centering}}
  \toprule[1pt] 
   & BLEU & Rouge-L & Dist-1 & Dist-2\\
   \hline
  \multicolumn{1}{m{1.5cm}|}{BM25} & 9.94 & 19.47 & 5.03$\%$ & 28.89$\%$ \\\hline
  \multicolumn{1}{m{1.5cm}|}{BERT-Retrieval} & 10.27 & 19.90 & \textbf{5.23$\%$} & \textbf{30.85$\%$} \\ \midrule[1pt]
  \multicolumn{1}{m{1.5cm}|}{Vanilla Seq2Seq} & 9.02 & 17.11 & 1.49$\%$ & 4.25$\%$ \\\hline
  \multicolumn{1}{m{1.5cm}|}{Seq2Seq-Attention} & 14.15 & 22.17 & 1.79$\%$ & 6.31$\%$ \\\hline
  \multicolumn{1}{m{1.5cm}|}{Seq2Seq-Copy} & \textbf{14.27} & \textbf{23.62} & 1.79$\%$ & 6.14$\%$ \\
  \bottomrule[1pt]
\end{tabular}
\caption{Automatic evaluation results. Dist-1/2 stands for Distinct-1/2.}
\label{tab:automatic evaluation results}
\end{table}

\begin{table*}[htb]
\centering
\begin{tabular}{l|cccc}
  \toprule[1pt] 
   & Ground Truth & BERT-Retrieval & Seq2Seq-Copy\\\hline
  Response Diversity & 88.74$\%$ & \bf 93.63$\%$ & 28.08$\%$ \\ \hline 
  ``Yes'' & \bf 4.10$\%$ & 0.94$\%$ & 3.74$\%$ \\
  ``What else can I do for you?'' & 3.42$\%$ & 0.88$\%$ & \bf 24.48$\%$ \\
  ``Wait a moment, I'll check for you right away'' & 0.50$\%$ & 0.18$\%$ & 3.90$\%$ \\
  \bottomrule[1pt]
\end{tabular}
\caption{Response diversity statistics and percentage of generic responses.}
\label{tab:response statistics}
\end{table*}

\begin{table*}[htb]
\small
\centering
\begin{tabular}{l|p{14cm}}
  \toprule[1pt]
  \multicolumn{2}{c}{Example 1}\\ \hline
  $q_1$ & \begin{CJK*}{UTF8}{gbsn}你好\end{CJK*} (Hi) \\
  $r_1$ & \begin{CJK*}{UTF8}{gbsn}你好\end{CJK*} (Hi) \\ 
  $q_2$ &  \begin{CJK*}{UTF8}{gbsn}帮我查下这个商品\end{CJK*} (Please check this item for me) \\ 
  $r_2$ & \begin{CJK*}{UTF8}{gbsn}好的，请问有什么可以帮您\end{CJK*} (Ok, what can I do for you?) \\ 
  $Q$ & \begin{CJK*}{UTF8}{gbsn}我要换这个摄像机，我这个上面绑定的账号能不能换绑？\end{CJK*} (I want to change this camera, may I change the bound account on this either?) \\ \hline
  BERT-Retrieval & \begin{CJK*}{UTF8}{gbsn}这个摄像机没有储存卡就不能回放。\end{CJK*} (You can't play back the video on the camera if you don't have the SSD.)\\ \hline
  Seq2Seq-Copy & \begin{CJK*}{UTF8}{gbsn}可以的哦。\end{CJK*} (Yes, it's ok.)\\
  \midrule[1pt]
  \multicolumn{2}{c}{Example 2}\\ \hline
  $q_1$ & \begin{CJK*}{UTF8}{gbsn}电子发票报销方便吗？\end{CJK*} (Is it convenient to reimburse with electronic invoice?) \\ 
  $r_1$ & \begin{CJK*}{UTF8}{gbsn}方便的，电子发票也是一样的\end{CJK*} (Yes, electronic invoice is the same.) \\ 
  $q_2$ & \begin{CJK*}{UTF8}{gbsn}稍等我问问会计\end{CJK*} (Wait a moment for me to ask the accountant.) \\ 
  $r_2$ & \begin{CJK*}{UTF8}{gbsn}好的\end{CJK*} (No problem.) \\ 
  $Q$ & \begin{CJK*}{UTF8}{gbsn}那就开电子票吧\end{CJK*} (Ok, send me electronic invoice please.) \\ \hline
  BERT-Retrieval & \begin{CJK*}{UTF8}{gbsn}好的，请问您税号多少呢？\end{CJK*} (Ok, what's your tax number please?)\\ \hline
  Seq2Seq-Copy & \begin{CJK*}{UTF8}{gbsn}好的，请问还有其他可以帮到您的吗？\end{CJK*} (Ok, what else can I do for you?)\\
  \bottomrule[1pt]
\end{tabular}
\caption{Examples of case study.}
\label{tab:case_study}
\end{table*}

The results of automatic evaluation are shown in Table \ref{tab:automatic evaluation results}. To further study the responses given by these models, we conduct some statistical analysis on the result sets. In Table \ref{tab:response statistics}, we present the response diversity (the portion of unique responses) in the Ground Truth set, BERT-Retrieval result set, and Seq2Seq-Copy result set. The portions of top three most common responses in the result sets are also listed in the table. Our observations can be summarized are as follows:
\begin{enumerate}
\item For retrieval-based models, BERT-Retrieval outperforms BM25, which shows the strong ability of pre-trained model in semantic matching task. For generative models, Seq2Seq-Copy performs the best, which shows the effectiveness of using attention and copy mechanism.
\item The generative model has better performance in the \textbf{similarity}  metrics (BLEU and Rouge-L) with the ground truth. There are mainly two reasons. One reason is generative model can generate new answers while retrieval models are limited by the $I$-$R$ pair repository. The other possible reason is that the ground truth also has many general responses (shown in Table \ref{tab:response statistics}), while the retrieval model tends to give responses containing specific information which may not fit the context and are quite different from the ground truth.
\item The retrieval-based model performs much better in response diversity (Dist-1/2 in Table \ref{tab:automatic evaluation results} and the response \textbf{diversity} in Table \ref{tab:response statistics}). While the performance of the generative model is very poor since it prefers to generate similar generic responses repeatedly (shown in Table \ref{tab:response statistics}), which is the common disadvantage of the generative models \cite{li2016diversity}.
\end{enumerate}

\subsubsection{Case Study}
We also show two representative cases in Table \ref{tab:case_study} to illustrate the difference between the two approaches intuitively. The retrieval model fails in the first case because it gives wrong information (talking about the ``SSD'', not the account), and the generative model fails in the second case for giving a generic response rather than useful information. From the cases above, we can see that the retrieval model tends to give responses with specific information (like ``SSD'' and ``tax number''), while the generative model usually gives generic answers (such as ``yes'' or ``what else can I do for you'').

For the frequently asked questions like invoice-editing, order-cancellation, etc., the retrieval model can perform well. However, For some specific questions which may never appear in the repository, the retrieval model may give a wrong answer, even though they contain the correct entities. For generative models, the generated generic responses may be lack of information, but they rarely make mistakes. Even not that satisfied, the users can still accept the generic responses sometimes. In general, both retrieval and generative models above are still not good enough. This shows the task complexity in the JDDC corpus.

\section{Conclusions and Future Work}
\label{label:conclusion}
In this work, we construct the Chinese JDDC dataset which is large-scale, multi-turn and collected in real scenario. We contribute three high-quality human-annotated challenge sets for better evaluation. Over 200 intents are also labelled for each query in the dataset. Besides, We evaluate several mainstream models on this dataset. The experimental results indicate either retrieval or generative models still have a long way to go in order to solve the real scenario conversation problem. More in-depth researches on context modeling, controllable response generation, question and answering, and reinforcement learning are needed in the future. Moreover, we will enrich the dataset annotations (e.g., emotions, and external knowledge) from various aspects in future work.  Our dataset is available at: \url{http://jddc.jd.com/auth_environment}, and we hope it can serve as an effective testbed and benefit future research in dialogue systems.

\section{Acknowledgements}
This work is partially supported by Beijing Academy of Artificial Intelligence (BAAI).

\section{Bibliographical References}
\label{main:ref}

\bibliographystyle{lrec}
\bibliography{JDDCDataset}


\end{document}